\documentclass[conference]{IEEEtran}
\usepackage{tikz}                                                                                                                                                                                    
\usetikzlibrary{arrows,shapes}

\usepackage{cite}

\usepackage{flushend}

\usepackage[cmex10]{amsmath}
\usepackage{amsfonts}
\usepackage{array}
\usepackage{mdwmath}
\usepackage{mdwtab}
\usepackage{fixltx2e}
\usepackage{stfloats}
\usepackage{url}
\newcommand\independent{\protect\mathpalette{\protect\independenT}{\perp}} 
\def\independenT#1#2{\mathrel{\rlap{$#1#2$}\mkern2mu{#1#2}}}

\begin{document}
% Please leave SVN version number $Revision: 406 $
% et ne pas oublier svn propset svn:keywords "Revision"  thisFile.tex:
\title{Strategic Planning in Air Traffic Control as a Multi-objective Stochastic Optimization Problem}

\author{
\IEEEauthorblockN{Ga\'etan Marceau}
\IEEEauthorblockA{Thales Air Systems and TAO, Rungis, France\\
Email: gaetan.marceaucaron@thalesgroup.com}
\and
\IEEEauthorblockN{ \ \ \ \ \ \ \ \ \ \ \ \ \ \ Pierre Sav{\'e}ant}
\IEEEauthorblockA{ \ \ \ \ \ \ \ \ \ \ \ \ \ \ \ \ Thales Research \& Technology, Palaiseau, France\\
\ \ \ \ \ \ \ \ \ \ \ \ \ \ \ \ \ Email: pierre.saveant@thalesgroup.com}
\and
\IEEEauthorblockN{ \ \ \ \ \ \ \ \ \ \ \ \ \ \ Marc Schoenauer}
\IEEEauthorblockA{ \ \ \ \ \ \ \ \ \ \ \ \ \ \ TAO, INRIA Saclay, France\\
 \ \ \ \ \ \ \ \ \ \ \ \ \ \ Email: marc.schoenauer@inria.fr}
}

%\author{\IEEEauthorblockN{Areski Hadjaz\IEEEauthorrefmark{1},
%Ga\'etan Marceau \IEEEauthorrefmark{2},
%Pierre Sav\'eant \IEEEauthorrefmark{3} and
%Marc Schoenauer\IEEEauthorrefmark{4}}
%\IEEEauthorblockA{\IEEEauthorrefmark{1}Thales Air Systems\\
%Rungis, France\\
%Email: areski.hadjaz@thalesgroup.com}
%\IEEEauthorblockA{\IEEEauthorrefmark{2}Thales Air Systems and Inria-Saclay\\
%Rungis, France\\
%Email: gaetan.marceau-caron@inria.fr}
%\IEEEauthorblockA{\IEEEauthorrefmark{3}Thales Research \& Technology\\
%Palaiseau, France\\
%Email:pierre.saveant@thalesgroup.com}
%\IEEEauthorblockA{\IEEEauthorrefmark{4}TAO, Inria Saclay\\
%Palaiseau, France\\
%Email:marc.schoenauer@inria.fr}}



\maketitle

\begin{abstract}
With the objective of handling the airspace sector congestion subject to continuously growing air traffic, we suggest to create a collaborative working plan during the strategic phase of air traffic control.
The plan obtained via a new decision-support tool presented in this article consists in a schedule for controllers, which specifies time of overflight on the different waypoints of the flight plans.
In order to do it, we believe that the decision-support tool shall model directly the uncertainty at a trajectory level in order to propagate the uncertainty to the sector level.
Then, the probability of congestion for any sector in the airspace can be computed.
Since air traffic regulations and sector congestion are antagonist, we designed and implemented a multi-objective optimization algorithm for determining the best trade-off between these two criteria. 
The solution comes up as a set of alternatives for the multi-sector planner where the severity of the congestion cost is adjustable.
In this paper, the Non-dominated Sorting Genetic Algorithm (NSGA-II) was used to solve an artificial benchmark problem involving 24 aircraft and 11 sectors, and is able to provide a good approximation of the Pareto front.
\end{abstract}



% \fi
\IEEEpeerreviewmaketitle

\section{Introduction}
\label{sec:intro}
Uncertainty in temporal projections is certainly a major difficulty in air traffic management.
It prevents the controllers from obtaining an accurate representation of the future airspace situation and it increases their workload when unpredictable events occur.

Today, the limits of ground trajectory prediction are known especially when the information concerning the aircraft state or the pilot intent is not available.  
Consequently, the air traffic controllers have to work with a bounded time horizon to ensure security separations. 
Beyond this horizon, the situation becomes unclear and anticipation is difficult, if at all possible. 
As a matter of fact, uncertainty has taken over. 
Nevertheless, with the sophistication of the flight management systems, one can expect that these limits will be pushed back.

This opens the way to new opportunities of planning during the so-called {\em strategic phase} of air traffic control. 
During this phase, flights can be scheduled at temporal points in order to minimize some space complexity critera and/or the delays of flight arrivals. 
The basic idea is to globally schedule beforehand all the flights in order to facilitate the work of the air traffic controllers in the tactical phase. 
Because the temporal horizon is then longer than in the tactical phase, the strategic phase must deal with the inherent uncertainties caused by the physical constraints. 
These can have different causes, from wind variation, to conflict resolutions to, more importantly as far as security is concerned, requested avoidance of an hazardous weather phenomenon area. 
Hence, it is mandatory for any automated system that is designed to optimize the scheduling of the flights in some airspace area to be aware of the uncertainties in order to take better decisions.

Today, any important textbook in optimal control, as \cite{Bertsekas:2000:DPO:517430}, or artificial intelligence, as \cite{DBLP:books/daglib/0023820} and \cite{Koller+Friedman:09} addresses the problem of modeling the uncertainties of the system at hand, so that the decisions are more robust to the possible perturbations. 
Moreover, the decisions should not be too conservative, as this could bring the system to a suboptimal behavior. 
In any case, uncertainty representation is relevant only if there is sufficient data to obtain reliable statistical results.

Fortunately, the amount of data recorded today in the operational centers is so huge that, on the opposite, any serious statistical study aiming at creating the uncertainty models requires an important amount of work in the area of data analysis. 
As in all Big Data applications, the main difficulty is to factorize/summarize/compress this information in a comprehensive and efficient way. 

This paper introduces an original methodology to tackle uncertainty regarding aircraft trajectories and airspace sector crossings. 
Probability theory is used as a formal framework to capture the essence of uncertainty in air traffic management. 
Then, the necessary tools for propagating the uncertainty temporally and spatially, through the waypoints of flight plans, are explained. 
From this propagation, we can infer the probability of sector congestion with a closed-form equation, avoiding costly Monte-Carlo simulations of the complete system that is usually the only way to numerically estimate uncertainties.
Then, the probabilistic model is used within an optimization algorithm in order to schedule all flights on the boundaries of the sectors in order to minimize the expected cumulated delays and the expected sector congestion.
To the best of our knowledge, the novelty of this work is to provide the inference mechanism to propagate the uncertainty from the trajectories to the sectors and to use the resulting probabilistic model as a black-box for multi-objective optimization.

The paper is organized as follows: Section \ref{sec:relatedWork} presents a literature survey on the formulations and the techniques used to solve the air traffic flow management problem.
Section \ref{section:Motivation} presents the motivation of the paper, while in Section \ref{section:mathematicalFormulation} the mathematical formulation of the uncertainty model and the optimization algorithm are defined. 
Section \ref{section:Experiments} presents some experimental results that further explain the model and the optimisation process.
Section \ref{section:Discussion} discusses on the possible extensions of the model and sketches further directions of research. 
Finally, Section \ref{section:Conclusion} concludes the paper and states some open questions.

\section{Related Work}
\label{sec:relatedWork}
The Operational Research community has studied many variants of the air traffic flow management problem since the beginning of the 90s. 
Along the years, the models have been refined in order to take into account new operational constraints. 
Today, we can distinguish two families of formulations: static approach and dynamic approach. 
The first one, also known as a single stage approach, is simpler in terms of formulation and computation, because its goal is to find an optimum once and for all.
The dynamic approach, or multi-stage approach, deals with uncertainty and incoming information about time estimates or unpredictable phenomenon. 

One of the first formulations of air traffic flow management is the ground holding problem, which minimizes the sum of airborne and ground delay costs when the demand for the runways exceeds the allowed capacities.
The decision variables are ground delays assigned to flights and the constraints are the capacities of the runways of the considered airports.
The first work under this formulation has been the Single Airport Ground Holding Problem \cite{Odoni1987}. 
After that, \cite{Richetta1994} has proposed a stochastic and dynamic version of the formulation and,  \cite{Mukherjee2007} has included the possibility to change marginal probabilities over a finite set of scenarios and to revise ground delays which were already assigned. 

Then, the Multi-Airport Ground Holding Problem was addressed by \cite{Vranas1994}.
This formulation addresses the problem by including a network of interconnected airports to propagate the delays, but does not take into account the sector capacities, rerouting and speed changes. 

The first two limitations were overcome with the model of \cite{Bertsimas1998}.
Also, this work has the merit to use realistic instances with several thousand flights.
To our knowledge, the most comprehensive formulation is the Air Traffic Flow Management Rerouting Problem \cite{LulliBertsimas2011} which integrates all phases of a flight, different costs for ground and air delays, rerouting, continued flights and cancellations. 
Instances of the size of the National Airspace of the United States were used to validate the approach.
Also, with the same mathematical framework, \cite{Agustn2012a}, \cite{Agustn2012b} have formulated the problem in terms of routes instead of nodes. 
The latter includes a stochastic formulation with discrete probabilities associated to scenarios for the capacities of the sectors.
In the same manner, there is also the work of \cite{Clare:2012} which describes an optimization problem to minimize directly the probability of congestion of the sectors with the concept of chance constraint.

The works mentioned so far use binary integer programming.
These techniques are powerful enough to address large-scale scenarios.
Besides, other techniques were used to solve similar problems.
\cite{Oussedik1998} uses stochastic optimization methods for handling sector congestion with takeoff delays and rerouting.
Constraint programming was also used by \cite{Barnier2001} and \cite{Flener2007}. 
The former solves the slot allocation problem with sector capacity constraints and the former minimizes an air traffic complexity metric for multiple sectors.

More recently, a multi-objective optimization approach has been used in air traffic control by \cite{Flener2007} to minimize an aggregated complexity metric, designed and validated by Eurocontrol, over sectors. In this case, the dimensions of the multi-objective problem are the complexity for each sector and thus, they use the weighted sum as a scalarization function to aggregate the complexity over all sectors. 
In this case, the multi-objective problem is transformed into a mono-objective problem and the weights are used to generate different trade-offs between sectors.
Also, \cite{daniel:2005} use the multi-objective to model the tradeoff between sector congestion and delays. 
In this case, the objective function space is in two dimensions and the parameter space consists of the takeoff time of the flights and the chosen routes.
A multi-objective genetic algorithm is used to generate a pool of solutions with a diversity measure in order to distribute them uniformly on the Pareto front.

Besides, a study of the uncertainty was conducted by \cite{Gilbo11} with an analysis of the prediction error of the time of arrival of the aircraft.
The main hypothesis of the study is that the random variable of the prediction error follows a Gaussian distribution.
The parameters of this distribution were estimated from real data.
The mean error is fixed to zero and the standard deviation is 4 minutes for active flights and 15 minutes for proposed flights.

To the best of our knowledge, this article is the first to tackle the problem of optimization of the strategic phase of air traffic control with a probabilistic model used to monitor the flights in real-time and a multi-objective algorithm to find the adequate actions to respond to disruptions.

%In this work, we will use evolutionary algorithms for multiobjective optimization because of the size of the parameter space. The Non-dominated Sorting Genetic Algorithm by , the Indicator-Based Evolutionary Algorithm by , the Strength Pareto Evolutionary Algorithm and the Covariance Matrix Adaptation Evolutionary Strategies are the selected algorithms for our study.
%
%The purpose of this paper is to propose a research path toward such a system, sketching both the formal and the numerical aspects of the resulting problem.

\section{Motivation}
\label{section:Motivation}
The goal of this work is to advance the state-of-the-art in strategic planning for air traffic control for strategic planning (up to 2 hours). We propose an innovative approach that takes into account the uncertainties of the flights. Furthermore, this approach must be dynamic because of the sliding time horizon of 2 hours: compared to the static day schedule that has been built days in advance, as flights takeoff and fly, the estimates on their time of arrival on each waypoint has to be frequently updated (this can be done via a data-link transmission between ground control and the flight management system of the aircraft).
The solution proposed here is to build a plan that is as robust to perturbations as possible, and to refine it as soon as more precise estimates of flights arrival at waypoints can be gathered.
This paper is intended to introduce a methodology to achieve the making of such plans and their monitoring through multi-objective optimization.
Then, a decision support tool can be easily derived from the multi-objective optimization results.
%%% stopped here - Marc

We believe that the proposed solution can enhance the airspace situation in congested areas.
The identified problem in such areas is the limited capacity of the runways for arrival, which will impact all adjacent sectors. 
In order to help the controllers, the arrival manager position was created to handle flight sequencing for each runway.
The resulting sequence enables the reduction of the four-dimensional space problem into a simpler one restricted to the route and the time. 
In such a configuration, the flights are heavily structured and the complexity is lower than in free flights as shown by the complexity measure defined in \cite{PuechmorelD09}. 
In situation of heavy traffic, such structure shall be generalized globally. 
To do so, as in \cite{Flener2007}, structuring can be established through objectives on coordination points between the sectors, such as the letters of agreement. 
The main difference is the use of an optimization algorithm for determining the best trade-off between conflicting interests in order to minimize the delays incurred by regulations and congestion of sectors.
The main novelty in this paper is to consider the uncertainty at a trajectory level by the intermediate of these objectives, then to compute the uncertainty propagation into the sectors by using a closed-form equation and finally, to use the probability marginals in order to compute the expected cost of delays and the expected cost of congestions as a multi-objective optimization problem. 
Moreover, the method presented here has the particularity to include the intentions or objectives directly in the generated solutions.

Contrary to static and deterministic approaches, a dynamic and stochastic approach, like the one presented here, can generate a plan that is robust to changes as long as the uncertainty is well-estimated. 
The difference with robust approaches, which consider the worst-case scenario, is that the plan is not too much conservative. 
In effect, the probability that the worst-case arises is so low that generating the plan according to it will systematically lead to a suboptimal behavior. Instead, it is more interesting to consider the scenarios proportionally to their probability of occurrence.

Besides, a multi-objective optimization approach of the strategic planning in air traffic control is a way to gather multiple alternatives on the arrival schedules in the sectors, each corresponding to a tradeoff between minimizing the use of regulations and reducing the complexity. 
In this work, the complexity refers to the congestion of a sector, which depends on its capacity. 
Usually, this value is determined statistically with data from the real-world system and consequently, nominal capacities are averages with associated variances. 
Since it is a scalar value for the entire sector, it does not account of the geometries of the trajectories for particular configurations. 
As a matter of fact, increasing the number of flights by one in the sector might increase drastically the workload of the controllers depending on the current airspace. 
If this one is already structured, the increase will be small, but if the flights have many crossing trajectories or with many flight level changes, the disruption might be significant. 
Consequently, the capacity threshold is not sufficient alone to evaluate the impact of the number of flights on the workload of the controllers. 
For this reason, we consider to give alternatives to the controllers where the severity of the congestion cost is variable. 
Of course, there exists more sophisticated complexity measures, which depend on the trajectories as presented in \cite{NASA:2000} , but then, the quality of the measures themselves becomes heavily tied to the accuracy of the trajectory prediction. 
As stated by \cite{sridhar2008}, ground trajectory prediction has trouble with determining accurately the 4D trajectories over twenty minutes. 
For these reasons, we estimate the number of aircraft in the sectors along time by taking the uncertainties around the trajectories of every flight. 
This measure is simple to analyze and we can assess directly the impact of the uncertainty of the flights on the capacity prediction. 
Nevertheless, the multi-objective paradigm is sufficiently general to include other complexity measures if deemed necessary in later stages.

\subsection{Multi-objective Optimization}
Multi-objective optimization is concerned with optimization problems involving several contradictory objectives. 
Given two solutions A and B of such multi-objective problems, A is obviously to be preferred to B in the case where all objective values for A are better than those of B, one at least being strictly better: in such case, A is said to {\em Pareto-dominate} B (denoted $A \prec B$). However, Pareto-dominance is not a total order, and most solutions are not comparable for this relationship. The set of interest when facing a multi-objective problem is the so-called {\em Pareto set} of all solutions of the search space that are not dominated by any other solution: such non-dominated solutions are the best possible trade-offs between the contradictory objectives, i.e., there is no way to improve any of them on one objective without degrading it on at least another objective. In multi-objective optimization, the search space is generally called the {\em design space}, by contrast with the {\em objective space} (one coordinate per objective), where the {\em Pareto front} is the image of the  set of non-dominated 
solutions -- the Pareto set. 

One common approach to multi-objective optimization is the so-called {\em aggregation method}, in which the goal is to minimize a single objective, some weighted sum of all objectives. The main advantage of this approach is that any generic optimization algorithm can then be used to minimize this single objective. However, this approach also suffers several drawbacks: it requires some a priori knowledge of the trade-off the decision maker is willing to make between the different objectives (materialized by the weights of the weighted sum); in case of doubt, one run per expected trade-off is necessary. Also, only convex parts of the Pareto front can be reached by the aggregation method.

A completely different approach is to use some population-based search, and to somehow factorize the efforts by identifying the whole Pareto front at once. Evolutionary Algorithms (EAs) \cite{EibenSmith2003}, bio-inspired optimization algorithms crudely mimicking natural evolution by implementing stochastic optimization through 'natural selection' and 'blind variations' can easily be turned into multi-objective optimizers by replacing the 'natural selection', that favors the best value of the objective function, by some 'Pareto selection' based on the Pareto dominance relation. One important observation is that a secondary selection criterion is needed, because Pareto dominance is not a total order relation: some diversity criterion is generally used, ensuring a wide spread of the population over the Pareto front. The resulting algorithms, Multi-Objective Evolutionary Algorithms (MOEAs), have demonstrated their ability to do in a flexible and reliable way \cite{Deb-book,Coello-book01}. 

MOEAs inherit several important properties from EAs : they are black-box stochastic optimization algorithms, in that they do not require any information on the objective functions (e.g., they do not require convexity, derivability, \ldots); they are robust to noise, an important property when dealing with real-world problems; unfortunately, they also inherit some dark side of EAs, in that they usually require a large number of objective computation.

Several MOEAs have been proposed in the literature, based on different implementation of the Pareto dominance selection and the diversity criterion. In particular, many MOEAs use an archive of solutions, where they maintain the non-dominated solutions ever encountered during the search. Finally, MOEAs have been applied to many real-world problems in optimal control and optimal design finance and robotics \cite{Coello-book04}.

\section{Mathematical Formulation}
\label{section:mathematicalFormulation}

\begin{figure}
  \centerline{
    \begin{tikzpicture}
      [scale=0.6,every node/.style={circle,fill=green!20,scale=0.7}]
      \node (n11) at (0,2) {$T_1^1$};
      \node (n12) at (2,2)  {$T_2^1$};
      \node (n13) at (4,2) {$T_3^1$};
      \node (n14) at (6,2) {$T_4^1$};
      \node (n21) at (0,-2) {$T_1^2$};
      \node (n22) at (2,-2)  {$T_2^2$};
      \node (n23) at (4,-2) {$T_3^2$};
      \node (n24) at (6,-2) {$T_4^2$};
      \node (t11)[rectangle,fill=red!20] at (0,4) {$\gamma_1^1$};
      \node (t12)[rectangle,fill=red!20] at (2,4) {$\gamma_2^1$};
      \node (t13)[rectangle,fill=red!20] at (4,4) {$\gamma_3^1$};
      \node (t14)[rectangle,fill=red!20] at (6,4) {$\gamma_4^1$};
      \node (t21)[rectangle,fill=red!20] at (0,-4) {$\gamma_1^2$};
      \node (t22)[rectangle,fill=red!20] at (2,-4) {$\gamma_2^2$};
      \node (t23)[rectangle,fill=red!20] at (4,-4) {$\gamma_3^2$};
      \node (t24)[rectangle,fill=red!20] at (6,-4) {$\gamma_4^2$};
      \node (ns1)[diamond,fill=blue!20] at (1,0) {$S_1$};
      \node (ns2)[diamond,fill=blue!20] at (3,0) {$S_2$};
      \node (ns3)[diamond,fill=blue!20] at (5,0) {$S_3$};
      \draw[->] (t11) edge (n11);
      \draw[->] (t12) edge (n12);
      \draw[->] (t13) edge (n13);
      \draw[->] (t14) edge (n14);
      \draw[->] (t21) edge (n21);
      \draw[->] (t22) edge (n22);
      \draw[->] (t23) edge (n23);
      \draw[->] (t24) edge (n24);
      \draw[->] (n11) edge (n12);
      \draw[->] (n12) edge (n13);
      \draw[->] (n13) edge (n14);
      \draw[->] (n21) edge (n22);
      \draw[->] (n22) edge (n23);
      \draw[->] (n23) edge (n24);
      \draw[->] (n11) edge (ns1);
      \draw[->] (n21) edge (ns1);
      \draw[->] (n12) edge (ns1);
      \draw[->] (n22) edge (ns1);
      \draw[->] (n12) edge (ns2);
      \draw[->] (n13) edge (ns2);
      \draw[->] (n22) edge (ns2);
      \draw[->] (n23) edge (ns2);
      \draw[->] (n13) edge (ns3);
      \draw[->] (n23) edge (ns3);
      \draw[->] (n14) edge (ns3);
      \draw[->] (n24) edge (ns3);
    \end{tikzpicture}
  }
  \caption{\label{fig:bn}Bayesian Network for a flight plan}
\end{figure}
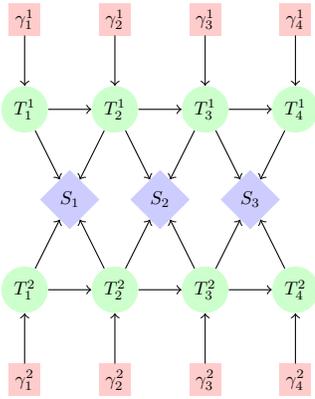

\subsection{Notations}
First, we need to define a probability space $(\Omega, \mathcal{T}, \Pr)$ where the sample space $\Omega=\mathbb{R}^+$ is an infinite temporal horizon beginning at the current moment, $\mathcal{T}$ is the set of events defined as the $\sigma$-algebra of the Borel sets on $\Omega$ and $\Pr$ is the Lebesgue measure on $\Omega$. 
Here, the events are the arrival time intervals at georeferenced points.
Let's consider a single flight plan in the set of the flight plans ($f \in \mathcal{F}$) defined as a mapping from a sequence of waypoints $X_1,\dots,X_n$ to a sequence of random variables $T_1^f, \dots, T_n^f$. 
We will use the standard definition  $\Pr \left[ T_i^f \in \Delta t \right] = \int_{\Delta t} p_i^f(t) \, dt$.
This simply refers to the probability for the flight $f$ to be over $X_i$ during the time interval $\Delta t$ and is obtained by integrating the density function $p_i^f$ over this interval.
Also, because we often consider the flight plan as an ordered sequence, we denote the joint probability density function from point $X_i$ to $X_j$ as
$p_{i:j}^f(t_{i:j})$, where $t_{i:j} \in \mathbb{R}^{j-i+1}$ is a vector with time values for every points of the sub-sequence.
Also, by convention, we denote the conditional probability: 
\begin{equation}
  \Pr \left[T_j^f \in \Delta t| T_i^f = t_i \right] = \int_{\Delta t} p_{j|i}^f(t|t_i) dt
\end{equation}
where $p_{i|j}^f(t|t_j)$ is the conditional density function associtad to the probability that the flight is over $X_j$ during the time interval $\Delta t$ given that the flight is over the point $X_i$ at time $t_i$.
In the following, we will drop the $f$ superscript when the formula applies to every flight.

\subsection{The Trajectory Model}
Let's define an uncertainty model for any trajectory.
To expose easily the concepts presented here, we rely on the graphical model on figure \ref{fig:bn}, 
namely a Bayesian Network, to represent the interactions between our random variables, represented with green circles.
In our graphical model, an arrow between $T_i$ and $T_{i+1}$ shows that the former influences the latter, or more precisely, that the two random variables are not independent.
The joint density function of $T_i$ and $T_{i+1}$ is:
\begin{eqnarray}
\label{eq:joint}
p_{i,i+1}(t_i,t_{i+1}) &=& p_{i+1|i}(t_{i+1}|t_i) \cdot p_i(t_i) \nonumber \\
&=& p_{i|i+1}(t_i|t_{i+1}) \cdot p_{i+1}(t_{i+1})
\end{eqnarray}
The first equality represents the propagation of the information in the same direction than the sequence of waypoints.
The second equation represents the propagation in reverse order, where information about $T_{i+1}$ is known before $T_i$.
From these two equations, the important Bayes' formula follows:
\begin{equation}
\label{eq:bayes}
p_{i|i+1}(t_i|t_{i+1}) = \frac{p_{i+1|i}(t_{i+1}|t_i) \cdot p_i(t_i)}{\int p_{i+1|i}(t_{i+1}|\tau) \cdot p_{i}(\tau) d\tau} \nonumber
\end{equation}
which enables backward inference.
As a first physical constraint, in order to respect the arrow of time along the sequence of waypoints of the flight plan, we first impose:
$$p_{i,j}(t_i,t_j) = 0, \; \mbox{ if } t_i \ge t_j, \; \forall j > i$$

Now, let's generalize the joint distribution for an arbitrary number of waypoints:
\begin{eqnarray}
\label{eq:fpModel}
p_{1:N}(t_{1:N}) &=& p_{N|1:N-1}(t_N|t_{1:N-1}) \cdot p_{1:N-1}(t_{1:N-1}) \nonumber \\
&=& p_{N|N-1}(t_N|t_{N-1}) \cdot p_{1:N-1}(t_{1:N-1}) \nonumber \\
&=& \dots \nonumber \\
&=& \prod_{i=2}^N p_{i|i-1}(t_i|t_{i-1}) p_1(t_1)
\end{eqnarray}
The first equality is obtained with the definition of the joint probability, the second equality requires the assumption of conditional independence $T_N \independent T_{1:N-2} | T_{N-1}$, also known as the Markov assumption. 
Then, the process is iterated for each $T_i$ from $N-1$ to $1$ in order to obtain the last equality.
Equation \ref{eq:fpModel} is the uncertainty model for the flight plan.
On the one hand, $p_1$ is the density function associated to the arrival of the flight in the airspace.
On the other hand, $p_{i|i-1}$ is the density function, which contains information on the intentions of the pilots to arrive at a point $i$ given the time of arrival on the previous point $i-1$. 
This formulation can be seen as an infinite number of ``what-if'' scenarios when temporal precision is not bounded. When discretizing the temporal space, we obtain a finite number of scenarios, but can still easily model the intention of increasing the speed if the flight was late on the previous point.

This model is more general than the one presented in \cite{atm:sid2012a}.
In this previous work, we have assumed the independence $D_{i,i+1} \independent T_i$ where $D_{i,i+1}$ is the random variable describing the duration to go  from $X_i$ to $X_{i+1}$.
Thus, we can use the fact that the sum of two independent random variables $(T_{i+1} = T_i + D_{i,i+1})$ is the convolution of the underlying probability density functions.
By defining the underlying probability density function of $D_{i,i+1}$ as $p_{i \rightarrow i+1}(t_{i+1} - t_i)$ and taking this result with eq. \ref{eq:joint}, we have:
\begin{eqnarray}
p_{i+1}(t_{i+1}) &=& \int p_{i \rightarrow i+1}(t_{i+1} - t_i) \cdot p_i(t_i) dt_i \nonumber \\
&=& \int p_{i+1|i}(t_{i+1}|t_i) \cdot p_i(t_i)dt_i
\end{eqnarray}
and so, in this case, $p_{i+1|i}(t_{i+1}|t_i) = p_{i \rightarrow i+1}(t_{i+1} - t_i)$.
The last equality is obtained by taking the marginal of $T_{i+1}$ of equation \ref{eq:joint}.
Notice that the independence assumption induces a mapping from the distributions in two dimensions into the distributions in one dimension. 
Effectively, this mapping will reduce the expressiveness of the model.
For this reason, in the following, this assumption of independence will {\bf not} be assumed.
Consequently, this complexity makes the underlying scheduling problem more difficult. 

\subsection{Uncertainty Around Sectors}
Here, we give the closed-form equation for computing the exact probability that a sector is congested. 
To do so, the probability that a flight is in a sector is necessary.
Let $S_{s,f}^t$ be the Bernoulli random variable that the flight $f$ is in the sector $s$ at time $t$ and $\overline{S_{s,f}^t}$ be its complementary.
Then, the probability to {\bf not} be in the sector during the time interval $t=[t_{min},t_{max}]$ is the probability to enter after $t_{max}$ or the probability to exit before $t_{min}$.
Because of the arrow of time constraint, these two events are mutually exclusive and one obtains:
\begin{eqnarray}
\label{eq:probSector}
  \Pr(\overline{S_{s,f}^t}) &=& \Pr(T_i > t_{max}) + \Pr(T_j \le t_{min}) \nonumber \\
  &=& \left[ 1 - \Pr(T_i \le t_{max}) \right] + \Pr(T_j \le t_{min}) \nonumber \\
  &=& 1 - F_i^f(t_{max}) + F_j^f(t_{min}) \nonumber \\
  \implies \Pr(S_{s,f}^t) &=& F_i^f(t_{max}) - F_j^f(t_{min})
\end{eqnarray}
where $F_i^f(t)$ is the cumulative density function denoting the probability that the flight $f$ has flown over $X_i$ by time $t$.
Now, inference on the presence of many flights in a given sector during an interval can be undertaken.
To do so, let $K_s^t$ be the random variable of the number of flights in the sector $s$ during the interval $t$.
Then, by using a multi-index notation, we have:
\begin{equation}
\label{eq:cong}
\Pr(K_s^t = n) = \sum_{|a|=n} \prod_{f \in \mathcal{F}} \Pr(S_{s,f}^t)^{a_f} \Pr(\overline{S_{s,f}^t})^{1-a_f}
\end{equation}
where $a = \left( a_1, a_2, \dots, a_{N_s^t} \right) \in \{0,1\}^{N_s^t}$, $|a| := a_1 + \dots + a_{N_s^t}$ and $N_s^t=| \left\{ i|\Pr(S_{i}^t) \ne 0 \right\}|$.
This refers to the Poisson Binomial distribution in statistics.
As an example, if we consider three flights, the equation becomes:
\begin{eqnarray}
\Pr(K_s^t = 1) &=& \Pr(S_{s,f_1}^t)\Pr(\overline{S_{s,f_2}^t})\Pr(\overline{S_{s,f_3}^t}) \nonumber \\ 
&+& \Pr(\overline{S_{s,f_1}^t})\Pr(S_{s,f_2}^t)\Pr(\overline{S_{s,f_3}^t}) \nonumber \\ 
&+& \Pr(\overline{S_{s,f_1}^t})\Pr(\overline{S_{s,f_2}^t})\Pr(S_{s,f_3}^t) \nonumber
\end{eqnarray}
As a first remark, the number of conjunctions (products) is determined by the number of combinations $\binom{N}{n}$ where $N$ is the total number of flights crossing the sector and $n$ is an arbitrary number of flights. 
Consequently, the associated computational burden attains its maximum value at $n = N/2$ and decreases when $n$ goes to $0$ or $N$.
As an example, the number of conjunctions is equal to $1.1826e+17$ when $N=60$ and $n=30$, which is intractable.
Nevertheless, one should be interested to know the probability of congestion and so, $n$ will take values from the capacity threshold to $N$.
In most cases, the difference between these shall be small. 
As an example, the number of conjunctions necessary to estimate the probability of congestion when $N= 60$ and the capacity is equal to 55 is 5985198, which is tractable.
Notice that we need to compute eq.\ref{eq:cong} for every values of $n \in \{55,\dots,60\}$.
That will give the probability at the specific time $t$.
So, we have one random variable per time, which is a stochastic process.
These are depicted with purple diamonds on the graphical model.

\subsection{Flight Intent}
At this point, one would like to manipulate flight intents more directly, i.e. for the generation of the conditional probabilities and during the optimization process. 
To do so, let $\gamma_{i+1}^f$ be the target time of arrival on the waypoint $X_{i+1}$ of the flight $f$.
Now, we make the strong assumption that the flights have a unique target time of arrival on each waypoint.
Then, the conditional probability can be parameterized as $p_{i+1|i}(t_{i+1}|t_i;\gamma_{i+1}^f)$.
An acceptable constraint on the space of possible conditional probabilities is to require an unimodal function where its maximum value is centered at the target value. 
Furthermore, we require that its support is bounded to denote the physical constraints of the aircraft, i.e. its flight envelope and its finite amount of fuel.
Good candidates for such properties are triangular and gamma probability density functions, already used in project management tools, like PERT, for characterizing the length of  a task in a scheduling problem.
Then, as depicted on the graphical model, the rectangular nodes act as interfaces between the optimization algorithm and the model for computing the expected cost functions.
This corresponds exactly to a black-box optimization approach.

\subsection{Optimization Formulation}
Now, we have all the elements to define our multi-objective optimization problem. 
Because of the stochastic context, one good way to define the cost functions is to use their expected values. 
For the expected cost of delays, let $\phi_f$ be the function associated to the cost of delays for the flight $f$. 
Then, the expected cost of delays for this flight is:
$$\mathbb{E}_{\phi_f}(T_{n^f}^f) = \int_{\Omega} \phi_f(t) \cdot p_{n^f}(t) dt < +\infty$$
where $n^f$ is the number of waypoints in the flight plan $f$ and so, $p_{n^f}$ refers to the marginal density function associated the arrival point $X_{n^f}$.
The inequality ensures that the cost function is bounded for the values in the support of the probability density function.
When aggregating these individual functions in order to obtain the associated objective function, one question that immediately arises is equity.
In this work, we define the same cost function for every flight and use the super-linear trick, from \cite{LulliBertsimas2011}, in order to penalize exponentially any delays. 
As a consequence, we avoid the case where a flight will be constantly penalized for the advantage of the other flights.
So, we use $\phi_f(t) = (t - A_f)_+^\beta$ where $A_f$ is the scheduled time of arrival of flight $f$, $\beta > 1$ is the super-linear coefficient and the plus symbol refers to the positive part.
One can also find other relevant cost functions without changing the optimization formulation.
In our work, we use: 
\begin{eqnarray} 
\label{eq:cost1}
\mathcal{C}_1(\gamma) &=& \sum_{f \in \mathcal{F}} \mathbb{E}_{\phi_f}(T_{n^f}^f;\gamma_{|f}) \nonumber \\ 
&=& \sum_{f \in \mathcal{F}} \left[ \int_{\Omega} (t - A_f)_+^2 \cdot p_{n^f}(t;\gamma_{|f}) dt \right]
\end{eqnarray}
as the first objective. Here, $\gamma_{|f}$ denotes the vector of objectives restricted to the ones concerning the flight $f$.
Notice that $p_{n^f}(\tau;\gamma_{|f})$ is the resulting marginal density function obtained from marginalizing the joint probability distribution obtained with eq. \ref{eq:joint} where all the components of $\gamma_{|f}$ are implied in the propagation of the uncertainty.

In the same manner, we define the congestion cost by $\psi_s(n,t)$ where $n$ is the number of flights at time $t$.
The expected cost of congestion is: 
$$\mathbb{E}_{\psi_s(\cdot,t)}(K_s^t) = \sum_{n=C_s}^{N_s} \psi_s(n,t) \cdot \Pr(K_s^t=n) < +\infty$$
where $C_s$ is the capacity of the sector $s$ and $N_s$ is the maximum number of flights that can be inside the sector at the same time with probability higher than 0.
Now, a difficulty arises when we aggregate along the time.
$\{S^t : t \in \Omega\}$ is a stochastic process where each random variable follows a Bernouilli distribution.
In order to make sense of the aggregation over $\Omega$, we need that the resultant of the integral of the expected cost of congestion is bounded.
As in optimal control, one way to overcome this difficulty is to use a factor, which decays proportionally with time.
The operational interpretation behind this solution is to give more importance to the sector that will be congested soon than those that will occur later for an equal probability. 
This can be done directly in $\psi_s(n,t)$.
Another solution is to bound the temporal horizon, but then, other difficulties arise with the normalization of the marginal probability over the chosen horizon.
At the current state of the research, we ignore these difficulties by choosing a temporal horizon sufficiently large to encompass the supports of the marginal distributions, denoted by $\bar{\Omega}$.
\begin{eqnarray}
\label{eq:cost2}
\mathcal{C}_2(\gamma) &=& \int_{\bar{\Omega}} \sum_{s \in \mathcal{S}} \mathbb{E}_{\psi_s (\cdot ,t)}(K_s^t;\gamma) dt \nonumber \\ 
&=& \int_{\bar{\Omega}} \sum_{s \in \mathcal{S}} \sum_{n=C_s+1}^{N_s} (n - C_s)_+^2 \cdot \Pr(K_s^t=n;\gamma) dt \nonumber
\end{eqnarray}
Again, $\Pr(K_s^t=n;\gamma)$ is the resulting probability distribution of the inference done with equation \ref{eq:cong}, which depends on the objectives $\gamma$. 

$\mathcal{C}_1$ and $\mathcal{C}_2$ are the two criteria of our bi-objective optimization problem.
Let $\mathcal{D} \subseteq \mathbb{R}^n$ be the decision space and $f:\mathcal{D} \rightarrow \mathbb{R}^2$ be the vector-valued cost function.
Let $\gamma \in \mathcal{D}$ be a point in our decision space, each dimension of $f(\gamma) = (\mathcal{C}_1(\gamma),\mathcal{C}_2(\gamma))$ denotes a cost associated to the decision.
In our problem, the two costs are antagonist, i.e. reducing the delays will induces more flights in the airspace and, as a consequence, will increase the congestion probability.
This idea is captured by the relation of strict Pareto dominance.
Let $\gamma_1,\gamma_2 \in \mathcal{D}$ be two decisions, then $\gamma_1$ strictly dominates $\gamma_2$, denoted by $\gamma_1 \prec \gamma_2$, iff $\mathcal{C}_i(\gamma_1) \le \mathcal{C}_i(\gamma_2), \; \forall i \in \left\{ 1,2 \right\}$ and $\exists j \in \left\{ 1,2 \right\} \;|\;  \mathcal{C}_j(\gamma_1) < \mathcal{C}_j(\gamma_2)$.

\subsection{Constraint}
From the optimization algorithm point of view, the objectives shall be bounded with the flight envelope. 
These bounds are hard constraints, which cannot be violated in order to find better solutions and define feasible intervals. 
There is a distinction between feasible intervals and the supports of the marginal distributions, since we can reasonably assign a probability zero to a point of the feasible interval. 
So, to be consistent, the support of the marginals must be subsets of the feasible intervals.
Consequently, we can only consider box constraints $\gamma_i^f \in \left[\underline{\gamma_i^f} , \overline{\gamma_i^f} \right]$, which are easily taken into account in evolutionary algorithms in general.
As an example, for the departure marginal distribution, if we assume that the flights respect their CFMU departure slot, which is a specificity to the European airspace, then the box constraint is defined by $\left[ \gamma_0^f -5, \gamma_0^f + 10 \right]$.
For en-route flight, we can propagate the hard constraints by assuming, as in \cite{Flener2007}, that a flight can have a maximum of speed up of 1 minute per 20 minutes and a maximum slow down rate of 2 minutes per 20 minutes. 
So, we have the interval recursive form:
\begin{equation}
\label{eq:recursive}
 \gamma_i^f \in \left[ \underline{\gamma_{i-1}^f} + \alpha \cdot D_{i,i+1}, \overline{\gamma_{i-1}^f} + \beta \cdot D_{i,i+1} \right]
 \end{equation}
 where $\alpha = 0.9$ and $\beta=1.05$ in this setting.

\subsection{Algorithm}
This section introduces the evolutionary optimization algorithm that has been used to validate numerically the theoretical framework described above, namely the Non-dominated Sorting Genetic Algorithm (NSGA-II)  \cite{Deb00afast}. 

NSGA-II uses a fast non-dominated sorting approach, with complexity $\mathcal{O}(MN^2)$, an elitism approach, and uses the so-called crowding distance as a secondary diversity criterion, that requires no additional parameter.  
All design variables here are continuous, and the variation operators have been chosen accordingly, as in \cite{Deb00afast}. More precisely, all experiments use the simulated binary crossover operator (SBX) and the polynomial mutation, which can handle directly the box constraints.
The other parameters of the algorithm are the population size, which is equal to the archive size, the probability of crossover, the probability of mutation, the dispersion coefficients, and the number of generations before termination.

NSGA-II has been chosen here for its long record of successes, demonstrating its robustness to find a good approximation of the Pareto front. Further work will investigate whether other MOEAs outperform NSGA-II on this problem: a good candidate is MO-CMA-ES \cite{igel2007ecj}, that has excellent performances on continuous problems.

\section{Experiments}
\label{section:Experiments}
In this section, the experiment on the multi-objective optimization on a probabilistic model of the trajectories and the sectors congestion is described, then the results on a simple instance are detailed. 
The first goal of these experiments is to numerically validate the theoretical model defined above, and to assess the propagation of the uncertainty from the trajectories to the sector congestion. 
The second goal is to assess whether NSGA-II can actually  solve the multi-objective optimization problem.

\subsection{Assumptions}
First of all, inside the probabilistic model, we need to discretize the temporal horizon in order to compute numerically the integrals.
We choose a time step of one minute because we believe that it is under the order of magnitude of the precision in the real-world.
This choice affects the accuracy of the evaluation of the uncertainty and therefore, is an internal parameter for the probabilistic model and can be used to control the tradeoff between accuracy and computational burden.
This parameter is completely hidden from the optimization algorithm because of the black-box optimization approach.
For this reason, we can use a continuous domain for the decision variables of the optimization problem that will permit any arbitrary small value for this parameter.
Thereafter, we assume that the feasible interval length for the first point is from 5 minutes to 60 minutes and the probability support length of the entry time is 15 minutes.
For the next points, we consider solely flights with constant level flight above FL300 with a true airspeed of 460 knots (MACH 0.78) and so, the probability support length here is smaller, i.e. it is fixed to 8 minutes.
As a matter of fact, the phases with flight level changes induce more uncertainty than a steady state flight and so, the propagation of the uncertainty that we will observe here will contain less uncertainty than in the real system.
With the coefficients in the recursive form defined at eq. \ref{eq:recursive}, it means that, for a constant distance, the flight can decrease its speed to 418 knots (MACH 0.72) and increase it to over 484 knots (MACH 0.82). 
Consequently, in this setting, we consider that the main source of uncertainty is caused by the wind.

We fix the average time in a sector to 10 minutes and we use a triangular distribution for every waypoint.
This should give enough opportunities to reduce the congestion probabilities.
From this distribution, we can generate all the duration between the waypoints of the flight plans.
Also, we need to define the conditional probability for every flight at every points, which require a lot of information.
Instead, an interesting way to do it is to assume a nominal policy for each flight.
The chosen nominal policy consists in trying to maximize the probability to arrive at the next point at the given objective. 
A way to encode the policy is to simply take the feasible intervals and to map a triangular distribution over it where the objective is the mode of the distribution.

%\twocolumn{
\begin{figure}
\centering
\includegraphics[scale=0.5]{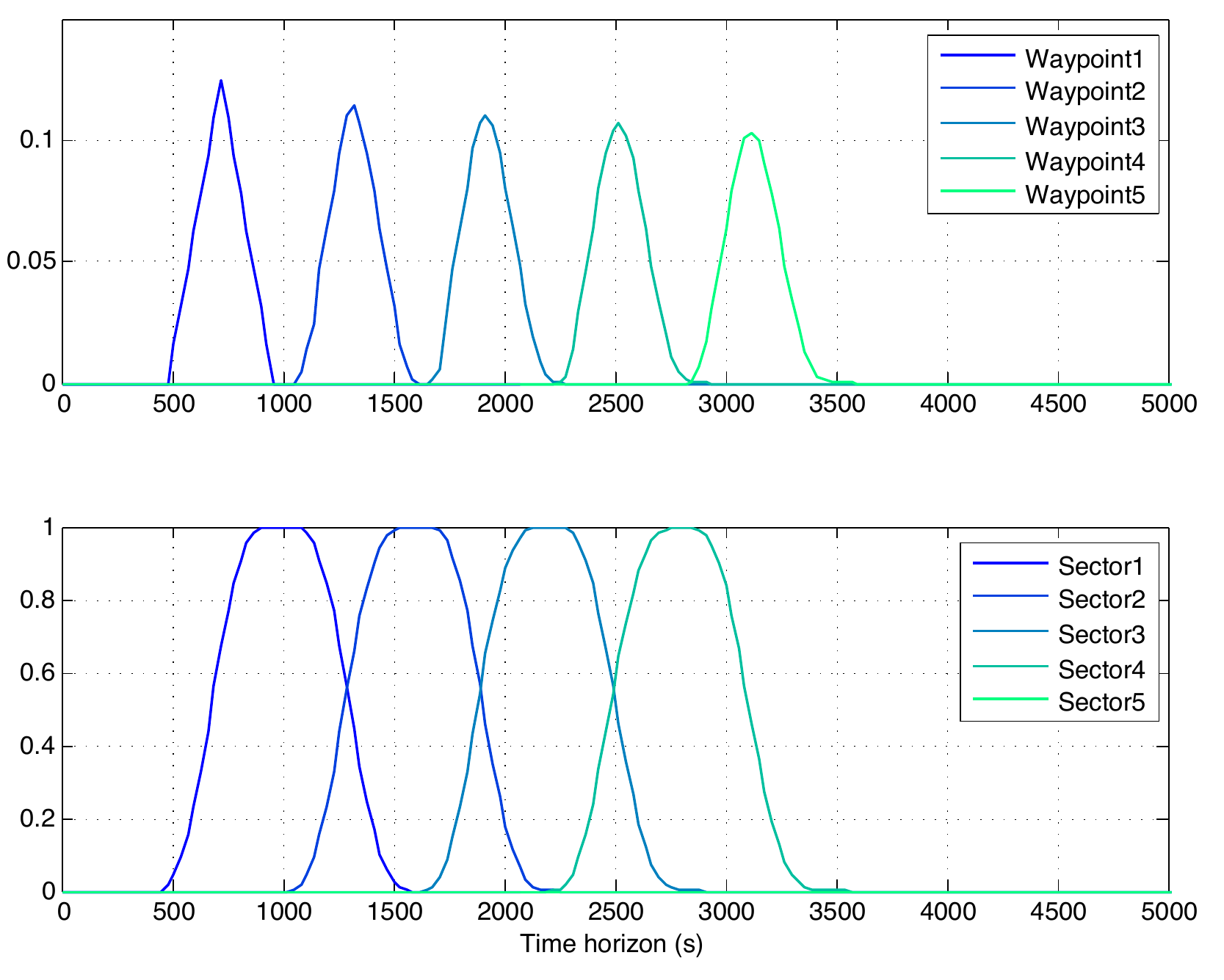}
  \caption{\label{fig:marginals} Uncertainty for one flight}
\end{figure}
%}

\subsection{Analysis}
For the first part of the analysis, we use a simple instance defined as a binary tree with 12 leaf nodes.
Two flights arrive at the same leaf with a gap of 2 minutes.
Every flight has the same uncertainty and their arrival node is the root of the tree.
Notice that the flights arriving at the first eight nodes will have 4 waypoints and those at the last 4 leaf nodes will have only 3 waypoints.
The sectors cover pair of adjacent branches with a capacity equals to the number of aircraft flying through the sector minus one. 
This gives 24 flights and 11 sectors. 
This can represent a toy example for an approach phase where the algorithm must sequence the flights during their convergence toward a single runway.
%Every distances on the branches of the tree are randomly generated following the target speed and the uniform distribution described in the assumptions. 
%This will avoid symmetrical solutions that rarely occur in the real world.

So, the first step consists in computing the marginal probabilities over the waypoints, as depicted on the upper part of figure \ref{fig:marginals}.
In this figure, we can see that the chosen policy tends to smooth the probability mass on the objectives.
As a matter of fact, for any possible value on the previous point, the flight management system will adapt its behavior to attain the next point at the given objective.
As the lookahead increases, the aircraft can rectify the bias at the departure and the asymmetry of the marginals disappears.
Also, the marginals seem the converge towards a distribution similar to a Gaussian distribution, but we do not know if it will happen for other suitable distributions.
Such considerations must be taken into account when choosing parametric distributions.
%One interesting result is the exponential decays on the tails of the distribution, which become more important with the number of waypoints. 
%This is mainly du to the products of the triangular function of the marginals by the triangular functions of the conditional probabilities.
%It seems that the resulting marginals underestimate the real probability, surely because the modes of the triangular distribution are always located on the objectives and that the possible variations of time are small. 

\begin{figure}
\centering
\includegraphics[scale=0.45]{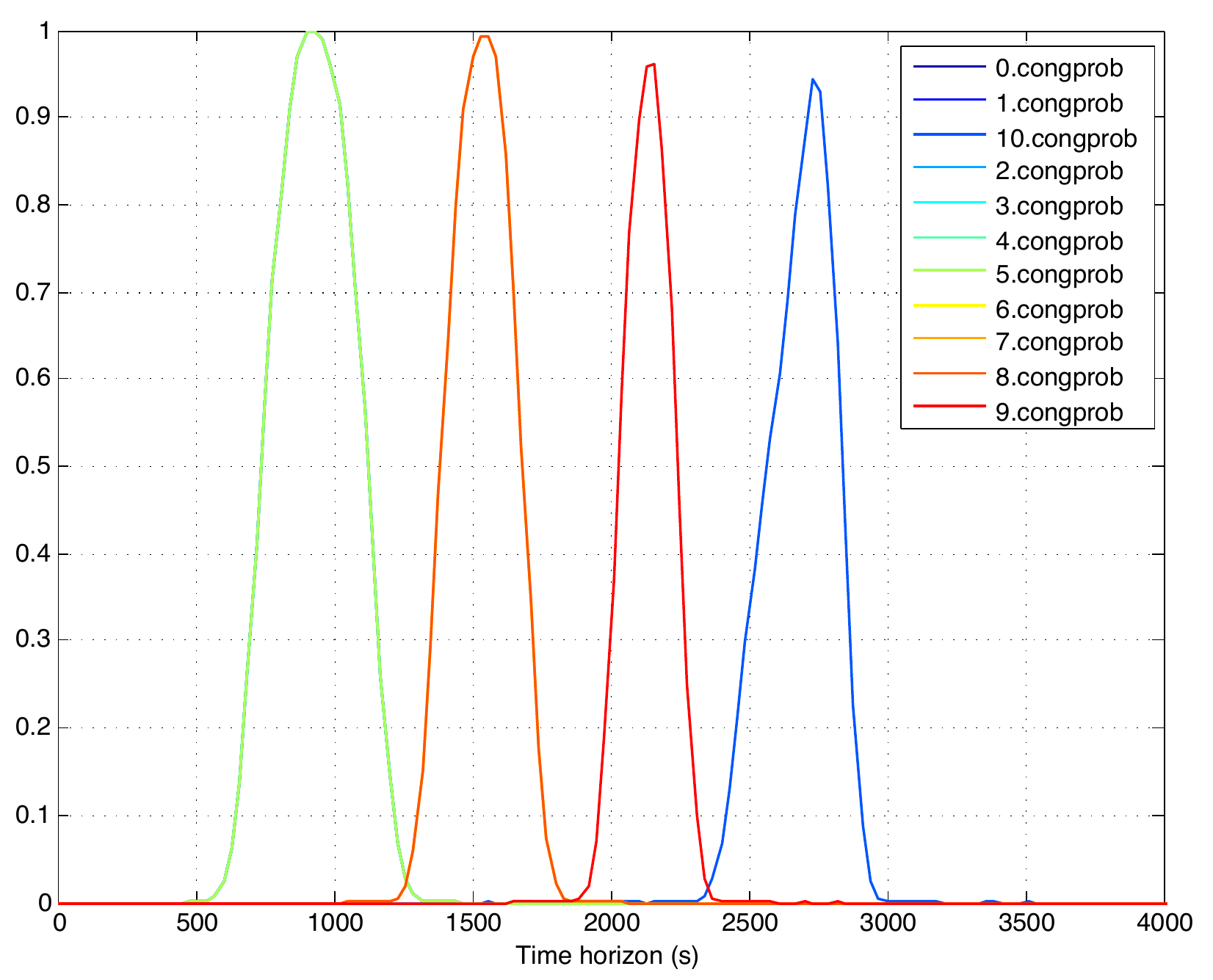}
  \caption{\label{fig:probCongestion}Probability of congestion for all sectors}
\end{figure}

The next step is the computation of the probability for a flight to be in a sector along the time. 
This is done with the equation \ref{eq:probSector} and the lower part of figure \ref{fig:marginals} shows the results.
As expected, for the sector probability, we can see that the overlap between sectors corresponds to the probability of overflight of the boundary points.
Also, their shapes look like Gaussian, but at this point of the study, we do not know if it is due to the choice of triangular distribution.

Due to the peaks of the marginal distributions, the variations of the probability to be in a sector are important during the entrance and the exit. 
This is a direct consequence of the assumptions and so; the uncertainty propagation is consistent with them. 
More generally, we can expect that the probability to be in a sector can be represented by a particular function.
The probability increases when the aircraft approaches the sector and once the flight is inside the sector, the probability should not decease until the exit.  

Thereafter, we need to compute the probability that there will be $n$ flights inside the sector at a given time with eq. \ref{eq:cong}.
Figure \ref{fig:probCongestion} shows the overlap of the probability of congestion for the 11 sectors.
We set the capacity of the sectors to one value under the number of flights that crossed it.
The first 6 sectors overlap on the first peak; the next 3 sectors overlap on the second followed by sector 9 and 10.
As expected, we see that the probability of congestion decreases with time, since uncertainty increases in the trajectories.
This can be viewed by comparing sector 6-7-8 with sector 9.
Also, we can see that the congestion probability of sector 10 is not symmetric since it receives the flights 15 to 23 before the others.
%Because we have two flights with the same parameters at the leaf nodes of the instances, we use the same color to denote the pairs with the same parameters.
%Since the capacity of the sector 6 is 6, we can see that the probability of congestion increases rapidly when flights 1, 2, 3, 5, 6 and 7 are in the sector during the time interval $\left[ 30,50 \right]$.

When all these distributions are known, the probabilistic model can compute the expected cost of delays and expected cost of congestion with eq. \ref{eq:cost1} and eq. \ref{eq:cost2} respectively.
One way to understand the cost functions from a computational point of view is to sum for all possible timestamp on the temporal horizon, the cost function at a given timestamp multiplied by the probability at this timestamp. 
Consequently, minimizing the probability of congestion for every timestamp will effectively minimize the expected cost function of congestion.

\begin{figure}
\centering
\includegraphics[scale=0.29]{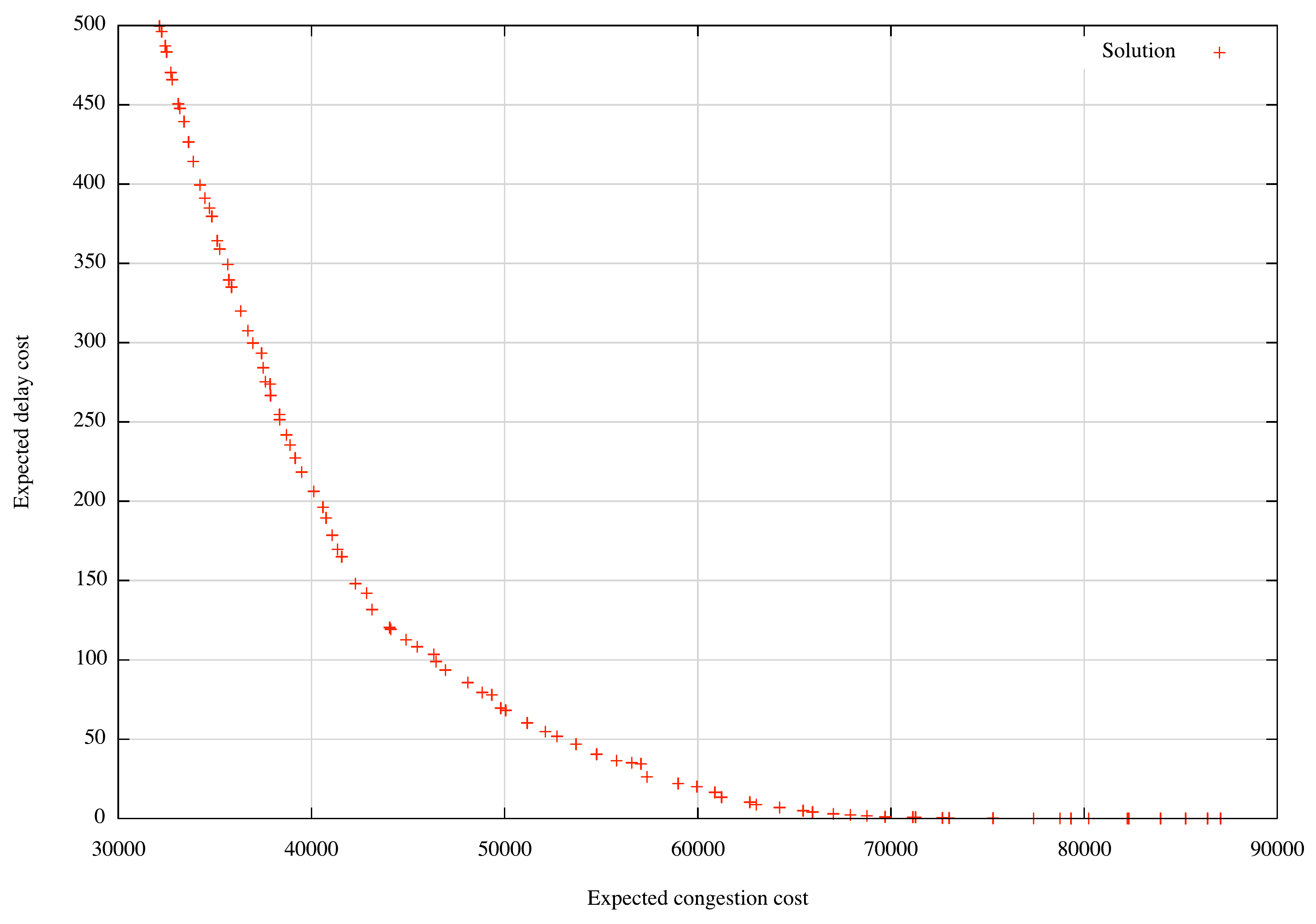}
  \caption{\label{fig:Pareto}Pareto front with 100 different solutions}
\end{figure}

Finally, when the cost functions are known, one can optimize by given different objectives to the flights. 
Figure \ref{fig:Pareto} shows the Pareto front for 100 different solutions.
Every solution is a complete schedule in the decision space and so, the Pareto front shows the trade-off between minimizing both the delays and the congestion.
If the decision maker wants to minimize the chance that delaying the flights congests the sectors, he shall select a solution toward the lower right corner of the graph. 
Otherwise, if she/he believes that the controllers can manage more flights, he can choose a schedule in the top left corner of the graph, resulting in much less delays, at the price of a higher congestion in some sectors.

\section{Discussion}
\label{section:Discussion}
In this work, we have validated the proposed theoretical framework on a single instance, assessing both the uncertainty model and the optimization process. Further experiments are mandatory to draw any firm conclusion regarding either the model or the optimization algorithm, and these experiments must involve several other instances 

For the probabilistic model, because the uncertainty on sectors are evaluated using the closed-form equations, the expected cost functions can be computed exactly for any instance \ldots assuming infinite computing capabilities. 
Regarding the optimization algorithm, there is very little hope to ever formally prove its convergence. 
Hence methods from experimental sciences must be used here. Statistics over many random instances of the size of a real operational context are the way to go, assessing how often and in which contexts the method can fail. 
This validation method will also provide insights on the actual computation burden that is required for large-scale instances. 

This opens the way to research on approximation of the probability of congestion by using sampling techniques e.g. Markov Chain Monte Carlo. 
Since we are in an online context, particle filters could also be used at the trajectory level in order to estimate the probability to be at the next points and therefore, estimate the probability of congestion. 
Since we have the closed-form, we can measure the accuracy of the estimation filters, which shall be much faster, with the exact method.

The choice of NSGA-II was justified by the fact that we do not fully understand all the properties underlying the probabilistic model. 
Instead of making false assumptions, we have chosen to use a robust and general optimization algorithm. 
Nevertheless, it is important to compare NSGA-II with more recent algorithms (e.g., as already mentioned, MO-CMA-ES \cite{igel2007ecj}). 
Therefore, an extensive statistical study is needed, in order to find the most adapted algorithm for this kind of problem.

Moreover, the formulation of the strategic planning problem shall also be extended to include more operational constraints and their effects on the algorithms. 
Also, experiments and data mining shall be done in an operational context in order to model accurately the underlying uncertainty.
Moreover, the uncertainty could represent the expected errors of the trajectory prediction used in the tactical phase. 
This would be create a smooth transition between the two phases.

Finally, we have mentioned the use of stochastic processes for the probability for the flight to be in a sector and for the probability of congestion. We believe that there is interesting questions that can be addressed with the novel techniques from this domain, notably on the analysis of the results generated by our method.

\section{Conclusion}
\label{section:Conclusion}
This article has introduced a probabilistic model to manage the propagation of the uncertainty of plane schedules along the waypoints of the flight plans. The prerequisites of the model are the marginal of the initial arrival in the airspace (takeoff time), conditional probabilities represented by the policy of the flight management system when trying to stick to the target schedule, and the potential external disruptions. Thereafter, the equation for computing the probability for a flight to be in a sector can be computed. From there on, the most important result in this work, the closed-form equation to compute the probability of congestion, can be derived. 

Then, some general formulations for the expected cost of delays and the expected cost of congestion were given. We used a well-known trick to ensure equity that was naturally integrated in the model. Finally, because the congestion measure is clearly not the only criterion that should be used to decide for a schedule, the well-known multi-objective algorithm NSGA-II was proposed to solve the bi-objective problem of minimizing both the congestion and the cumulated delays of the flights, i.e., to approximate the non-dominated solutions of the Pareto front. These solutions are then proposed as alternatives for the decision maker, namely here the multi-sector planner.

Furthermore, in order to illustrate how the theoretical model can be useful in practice, we presented some results on an instance with 24 flights and 11 sectors. 
The results were analyzed to discover the consequences of some previous assumptions. 
One is the choice of triangular distributions, commonly used in project management, which seems to return results coherent with our intuition.
On-going and further work will of course investigate other MOEAs to replace NSGA-II, and, more importantly, several different instances. One crucial issue is how well (or bad) this algorithm scales with the problem complexity (number of flights and number of sectors).
Nevertheless, we are confident that further studies will demonstrate the robustness of the proposed approach of using multi-objective evolutionary algorithms to solve the stochastic optimization problem of air traffic en-route planning and control.

\section{Acknowledgments}
Ga{\'e}tan Marceau is funded by the scholarship CIFRE 710/2012 established between Thales Air Systems and the TAO project-team at INRIA Saclay Ile-de-France, and the scholarship 167544/2012-2013 from the {\it Fonds de Recherche du Qu{\'e}bec - Nature et Technologies}.

\bibliographystyle{IEEEtran}
\bibliography{library}

\section{Author biographies}
{\bf Ga{\'e}tan Marceau} is a Ph.D. student in computer science in the TAO team at INRIA Saclay Ile-de-France, working in the ``Laboratoire de Recherche en Informatique'' of the University Paris-Sud. His thesis is also co-supervised and co-funded by Thales Air Systems France, a major actor in the development and deployment of operational Air Traffic Control Centers.

{\bf Pierre Sav{\'e}ant} is a research engineer at Thales Research \& Technology France.
He obtained his Ph.D. in Artificial Intelligence from the P. \& M. Curie University in 1990.
He has been involved in modeling and solving combinatorial optimization problems for civil and defense applications for more than 25 years. His current research interests include automated planning and evolutionary computing.

{\bf Marc Schoenauer} is Senior Researcher with INRIA Saclay Ile-de-France, co-head of the TAO team. He has been working in the field of Evolutionary Computation since the early 90s, is author of more than 120 papers in journals and major conferences of this field. He has been (co-)advisor of 28 PhD students. 
He is member of the Executive Board of SIGEVO, the ACM Special Interest Group for Evolutionary Computation. He has been Editor in Chief of Evolutionary Computation Journal (2002-2010), and serves on the Editorial Board of the main journals and on the Program Committees of all major conferences in the field of Evolutionary Computation.

\end{document}